\documentclass[9pt, conference]{IEEEtran}
\IEEEoverridecommandlockouts

\makeatletter
\def\ps@IEEEtitlepagestyle{%
  \def\@oddfoot{\mycopyrightnotice}%
  \def\@oddhead{\hbox{}\@IEEEheaderstyle\leftmark\hfil\thepage}\relax
  \def\@evenhead{\@IEEEheaderstyle\thepage\hfil\leftmark\hbox{}}\relax
  \def\@evenfoot{}%
}
\def\mycopyrightnotice{%
  \begin{minipage}{\textwidth}
  \scriptsize
  \copyright~2023 IEEE. Personal use of this material is permitted. Permission from IEEE must be obtained for all other uses, in any current or future media, including reprinting/republishing this material for advertising or promotional purposes, creating new collective works, for resale or redistribution to servers or lists, or reuse of any copyrighted component of this work in other works. 
  
  This work has been accepted at the Design Automation Conference (DAC’23).
  \end{minipage}
}
\makeatother

\usepackage{amsmath,amssymb,amsfonts}
\usepackage{graphicx}
\usepackage{textcomp}
\def\BibTeX{{\rm B\kern-.05em{\sc i\kern-.025em b}\kern-.08em
    T\kern-.1667em\lower.7ex\hbox{E}\kern-.125emX}}
\usepackage{etoolbox}
\makeatletter
\newcommand{\linebreakand}{%
  \end{@IEEEauthorhalign}
  \hfill\mbox{}\par
  \mbox{}\hfill\begin{@IEEEauthorhalign}
}
\patchcmd{\@maketitle}
  {\addvspace{0.5\baselineskip}\egroup}
  {\addvspace{-1.3\baselineskip}\egroup}
  {}
  {}
\makeatother

\usepackage{cite}
\usepackage{amsmath,amssymb,amsfonts}
\usepackage{graphicx}
\usepackage{textcomp}
\usepackage{bbding}
\usepackage{pifont}
\usepackage{multicol}

\usepackage{booktabs}
\usepackage[binary-units=true]{siunitx}
\usepackage{array,multirow}
\usepackage{hyperref}
\usepackage{amsmath}
\usepackage{enumitem}
\usepackage{algorithm}
\usepackage{graphicx}
\usepackage{float}
\usepackage[symbol, hang,flushmargin]{footmisc}
\usepackage{bm}

\usepackage{tablefootnote}

\usepackage[noend]{algpseudocode}

\newcommand{\figref}[1]{Figure~\ref{#1}}
\newcommand{\secref}[1]{Section~\ref{#1}}

\newcommand{\tabref}[1]{Table~\ref{#1}}

\newcolumntype{L}[1]{>{\raggedright\let\newline\\\arraybackslash\hspace{0pt}}m{#1}}
\newcolumntype{C}[1]{>{\centering\let\newline\\\arraybackslash\hspace{0pt}}m{#1}}
\newcolumntype{R}[1]{>{\raggedleft\let\newline\\\arraybackslash\hspace{0pt}}m{#1}}

\usepackage{soul}
\usepackage[dvipsnames]{xcolor}
\usepackage{xurl}

\definecolor{light_red}{RGB}{255, 204, 204}
\definecolor{crimson}{rgb}{0.86, 0.08, 0.24}    

\newif\ifcomment

\commentfalse

\ifcomment
\newcommand{\hongxiang}[1]{\sethlcolor{yellow}\hl{[Hongxiang: #1]}}
\newcommand{\thomas}[1]{\sethlcolor{magenta}\hl{[Thomas: #1]}}
\newcommand{\stelios}[1]{\sethlcolor{green}\hl{[Stelios: #1]}}
\newcommand{\alex}[1]{\sethlcolor{cyan}\hl{[Alex: #1]}}
\newcommand{\royson}[1]{\sethlcolor{purple}\hl{[Royson: #1]}}
\newcommand{\mohamed}[1]{\sethlcolor{orange}\hl{[Mohamed: #1]}}
\newcommand{\cut}[1]{\sethlcolor{light_red}\hl{[#1]}}

\else
\newcommand{\hongxiang}[1]{}
\newcommand{\thomas}[1]{}
\newcommand{\stelios}[1]{}
\newcommand{\alex}[1]{}
\newcommand{\royson}[1]{}
\newcommand{\mohamed}[1]{}
\newcommand{\cut}[1]{}

\fi

\begin{document}

% \title{Optimizing Multi-Exit Bayesian Neural Networks on FPGA \vspace{-5mm}}

\title{When Monte-Carlo Dropout Meets Multi-Exit: \\ Optimizing Bayesian Neural Networks on FPGA\vspace{-2.5mm}}

% \title{Automating FPGA-based Acceleration of \\ Multi-Exit Bayesian Neural Networks \vspace{-5mm}}

\author{
\IEEEauthorblockN{Hongxiang Fan\IEEEauthorrefmark{2}}\thanks{\IEEEauthorrefmark{2} Work was done while pursuing Ph.D. at Imperial College London.}
\IEEEauthorblockA{
% \textit{Department of Computing} \\
% \textit{Imperial College London}\\
% London, UK \\
\textit{Samsung AI Center \&} \\
\textit{University of Cambridge}\\
Cambridge, UK \\
hongxiangfan@ieee.org}
\and
\IEEEauthorblockN{Hao (Mark) Chen}
\IEEEauthorblockA{\textit{Department of Computing} \\
\textit{Imperial College London}\\
London, UK \\
hao.chen20@imperial.ac.uk}
\and
\IEEEauthorblockN{Liam Castelli}
\IEEEauthorblockA{\textit{Department of Computing} \\
\textit{Imperial College London}\\
London, UK \\
castelliliam@gmail.com}
\linebreakand
\IEEEauthorblockN{Zhiqiang Que}
\IEEEauthorblockA{\textit{Department of Computing} \\
\textit{Imperial College London}\\
London, UK \\
z.que@imperial.ac.uk}
\and
\IEEEauthorblockN{He Li}
\IEEEauthorblockA{\textit{Department of Electronics and Engineering} \\
\textit{Southeast University}\\
Nanjing, China \\
helix@seu.edu.cn}
\and
\IEEEauthorblockN{Kenneth Long}
\IEEEauthorblockA{\textit{Department of Physics} \\
\textit{Imperial College London}\\
London, UK \\
k.long@imperial.ac.uk}
\and
\IEEEauthorblockN{Wayne Luk}
\IEEEauthorblockA{\textit{Department of Computing} \\
\textit{Imperial College London}\\
London, UK \\
w.luk@imperial.ac.uk}
}

\maketitle

\begin{abstract}
Bayesian Neural Networks (BayesNNs) have demonstrated their capability of providing calibrated prediction for safety-critical applications such as medical imaging and autonomous driving.
However, the high algorithmic complexity and the poor hardware performance of BayesNNs hinder their deployment in real-life applications.
To bridge this gap,
this paper proposes a novel multi-exit Monte-Carlo Dropout (MCD)-based BayesNN that achieves well-calibrated predictions with low algorithmic complexity.
To further reduce the barrier to adopting BayesNNs,
we propose a transformation framework that can generate FPGA-based accelerators for multi-exit MCD-based BayesNNs.
Several novel optimization techniques are introduced to improve hardware performance.
Our experiments demonstrate that our auto-generated accelerator achieves higher energy efficiency than CPU, GPU, and other state-of-the-art hardware implementations.
Our code is publicly available at: \url{https://github.com/os-hxfan/BayesNN_FPGA.git}
% We intend to open-source our design after the paper acceptance.

% including spatial\&temporal mapping and algorithm\&hardware co-exploration.

% To reduce the barrier to the adoption of FPGA-based BayesNNs, 
% this paper proposes \textit{AutoBayes}, an end-to-end transformation tool that can automatically generate BayesNN hardware  accelerators based on high-level synthesis, given as input a non-Bayesian software model.
% A novel feature of \textit{AutoBayes} is its support for mapping multi-exit architectures into hardware. AutoBayes enables the first systematic study quantifying the benefits and the overheads of FPGA-based BayesNNs with respect
% to their non-Bayesian counterparts. The purpose is to help application builders to appreciate the pros and cons of deploying FPGA-based BayesNNs, so that they can be considered for adoption especially for safety-critical applications, for which the provision of uncertainty information is critical.
% Our experiments show that the overheads of FPGA-based BayesNN designs from \textit{AutoBayes} are often within a few percent of their non-Bayesian versions, and sometimes the FPGA-based BayesNNs can have improved accuracy compared with the non-Bayesian versions. 
% We intend to open source the \textit{AutoBayes} tool and to submit it for artifact evaluation.

\end{abstract}


\section{Introduction}

Deep neural networks (DNNs) have become a frontier of artificial intelligence, with a variety of applications in many domains including computer vision, natural language understanding, medical image processing and scientific data analysis~\cite{dong2021survey}. However, conventional deep neural networks have an important drawback: they behave like black boxes so can neither explain how the answers are obtained, nor provide an estimate of their confidence in their correctness.
Bayesian Neural Networks (BayesNNs)~\cite{neal1993bayesian} have been introduced to address the lack of confidence estimation of conventional deep neural networks, or non-Bayesian NNs (non-BayesNNs). The uncertainty-aware feature of BayesNNs makes them more resilient to risks caused by over-confident prediction of non-BayesNNs.

Nevertheless,
there are two major challenges in deploying BayesNNs in real-life applications.
First,
the high dimensionality of modern BayesNNs significantly increases their algorithmic complexity,
making exact Bayesian inference intractable.
Although various approximation approaches, such as variational inference~\cite{blundell2015weight} and Monte-Carlo Dropout (MCD)~\cite{gal2016dropout}, have been introduced to reduce computational overhead,
these approaches perform worse in terms of uncertainty quality and calibration ability~\cite{ovadia2019can} than traditional deep ensembles that consist of multiple DNNs.
Second,
even with approximations,
the computational and memory demands of BayesNNs are still much higher than those of non-BayesNNs due to Monte-Carlo (MC) sampling, hindering their deployment in demanding applications, especially those with real-time requirements.
While there is extensive research on hardware acceleration for deep learning algorithms, most existing efforts focus on domain-specific hardware~\cite{fowers2018configurable, chen2016eyeriss, fan2022adaptable} or design automation tools~\cite{zhang2018caffeine, fahim2021hls4ml} for non-BayesNNs such as convolutional NNs (CNNs)~\cite{krizhevsky2017imagenet, he2016deep} and long short-term memory (LSTM)~\cite{hochreiter1997long}.
Hence there is an urgent need for publicly accessible hardware acceleration for BayesNNs.

To address the first challenge,
this paper proposes a novel multi-exit MCD-based BayesNN.
Compared with traditional MCD-based BayesNNs,
our method is able to provide well-calibrated predictions.
Also,
% our multi-exit MCD-based BayesNN is flexible to generate arbitrary numbers of MC samples compared with conventional deep ensembles.
the proposed multi-exit MCD-based BayesNNs are
less computational and memory-intensive than conventional deep ensembles while possessing the flexibility to generate arbitrary numbers of MC samples.
To overcome the second challenge,
we propose to accelerate multi-exit MCD-based BayesNNs on FPGA.
A transformation framework is introduced to generate high-performance FPGA-based accelerators for multi-exit MCD-based BayesNNs.
With several novel optimizations such as spatial-temporal mapping and algorithm-hardware co-exploration,
the generated accelerators achieve higher energy efficiency than previous hardware implementations.
% Although FPGA-based BayesNNs are appealing~\cite{fan2022fpga,cai2018vibnn,fan2022accelerating}, their wide-spread adoption would not happen unless they are easy to access. 
% Therefore, this paper proposes an open-source transformation framework that automatical generates FPGA-based accelerators for multi-exit MCD-based BayesNNs.
% With several novel optimizations such as spatial\&temporal mapping and algorithm\&hardware co-exploration,
% the generated accelerators achieve higher energy efficiency than previous hardware designs.
Moreover, this paper provides the first systematic study quantifying the benefits and the overheads of accelerated BayesNNs against their non-BayesNN counterparts. It would be of interest to DNN application builders to understand the trade-offs in deploying FPGA designs of BayesNNs to replace those implementing non-BayesNNs. 
% To facilitate public access to our designs,
% we intend to open-source our code after the paper acceptance.

The contributions of this paper can be summarized as follows:
\begin{itemize}[leftmargin=*]

\item A novel multi-exit MCD-based BayesNN with better calibration ability than conventional MCD-based BayesNN,
and higher computational efficiency and flexibility over traditional deep ensembles.

\item A design framework for transforming non-BayesNN models to multi-exit BayesNN hardware accelerators with high hardware performance and energy efficiency.

\item Various optimization strategies including spatial-temporal mapping and algorithm-hardware co-exploration for performance improvement.

% \item A comprehensive evaluation of the proposed approach based on multiple models and datasets, demonstrating its effectiveness in improving both algorithm and hardware performance.

\end{itemize}

\section{Background and Related Work}\label{sec:background}
\subsection{Bayesian Neural Networks}\label{sec:bnns}

BayesNNs are able to achieve robustness against overfitting and to provide the estimation of their model uncertainty by means of Bayesian inference. 
Instead of capturing point-wise weight values like non-BayesNNs, BayesNNs are trained to learn the distribution of the weights. The Bayes rule is adopted in learning the distribution $p(w|D)$ for the weights $w$ with respect to training data $D$.
It is, however, computationally intractable to calculate the posterior distribution $p(w|D)$ analytically due to the high dimensionality of modern BayesNNs. 
To address this issue, various approximation methods have been introduced for BayesNNs~\cite{ovadia2019can}.
Among these approaches,
Monte Carlo Dropout (MCD)~\cite{gal2016dropout} is drawing attention as it provides an efficient way to estimate uncertainty, which is achieved by interpreting the dropout training of DNNs as approximate Bayesian inference for deep Gaussian processes.

MCD is implemented by applying a random filter-wise mask to the output feature maps of a layer $i$ with $F_{i}$ dimensional filters, which randomly drops out connections in a neural network. The values for the mask $M_{i}$ adopt a Bernoulli distribution $p(M_{i}| p_{i})$ with binary random variables (0 or 1) with probability $p_{i}$. 
The inference of MCD-based BayesNNs requires running multiple forward passes to generate different MC samples.
In other words, the uncertainty estimation and calibration are achieved by feeding the same input through a BayesCNN $S$ times, each time with a different set of sampled masks $M$.
Dropouts have been employed in non-BayesNNs, typically during training. In contrast, for BasyesNNs, MCD-based dropout takes place during training as well as during inference. 

\subsection{Multi-Exit Network}\label{sec:multiexit}
Deep ensembles involve combining the predictions from multiple
individual neural networks. They enable high prediction accuracy and high quality of calibration and uncertainty quantification~\cite{deepensembles}. However, training and using these networks can be prohibitively expensive.
An alternative approach is to use a multi-exit architecture. By introducing intermediary classifiers before the final exit, multiple predictions can be obtained in a single pass. Using an equally weighted ensemble of the predictions from each of these exits can be shown to achieve accurate uncertainty estimation \cite{ee}.

While some architectures like the Multi-Scale DenseNet~\cite{msdnet} have been specifically designed for multi-exit, the most common approach to developing these architectures is to use a known powerful backbone architecture like ResNet, and then to attach intermediary classifiers at particular points \cite{bidistillation, sdn}. These exit points can be selected in a variety of ways by floating-point operation (FLOP) thresholds or semantic groupings of convolutional layers \cite{sdn, bidistillation}. 

% Training these networks has also been the focus of significant research, as training procedures need to balance the performance of each network without compromising the overall final accuracy. For classification tasks, standard weighted cross-entropy loss functions have been outperformed by more complicated approaches like those that incorporate aspects of knowledge distillation \cite{bidistillation}.
% A comparison of distillation approaches for multi-exit architectures~\cite{bidistillation} shows that bidirectional distillation has the best classification accuracy for a variety of model types on multiple datasets. 

\subsection{Related Work}

Much research has taken place on deep neural networks and their applications~\cite{dong2021survey}, and the use of FPGAs to accelerate deep neural networks. Representative work in this area includes energy-efficient CNN acceleration~\cite{chen2016eyeriss} and FPGA-based real-time AI cloud services~\cite{fowers2018configurable}. 
There has also been significant research into design automation for deep neural networks. One example is the open source tool {\it hls4ml} supporting an automatic design flow involving high-level synthesis to promote low-power machine learning~\cite{fahim2021hls4ml}.
% , and the {\it Caffeine} approach with uniformed representation and acceleration for CNNs~\cite{zhang2018caffeine}.

FPGA-based acceleration of BayesNNs has been reported recently~\cite{wan2021shift}. One example of an early design is {\it Bynqnet}, an FPGA-based Bayesian neural network with quadratic activations for sampling-free uncertainty estimation~\cite{awano2020bynqnet}. 
Efficient FPGA implementations for 2D and 3D BayesNNs have been reported~\cite{fan2022fpga}.  
Another example is {\it VIBNN}, an FPGA-based accelerator that supports variational inference in BaynesNNs~\cite{cai2018vibnn}. 
% Fan et al. show the development of efficient FPGA implementation for 2D and 3D BayesNNs
There is also research on algorithmic and hardware optimizations of BayesNNs, exploiting their structured sparsity and redundant computations~\cite{fan2022accelerating}. 
In contrast to these approaches,
this work proposes to accelerate multi-exit MCD-based BayesNNs on FPGA, achieving high energy efficiency and well-calibrated predictions.
% high energy efficiency and quality of predictive uncertainty.
% However, there has not been a tool based on HLS for producing FPGA-based BayesNNs. \textit{AutoBayes} is intended to address this gap in design automation of FPGA implementations for BayesNNs.

\section{Multi-Exit Meets MCD}

As discussed in~\secref{sec:background}, both multi-exit and MCD-based approaches are able to generate calibrated predictions, 
but they also have clear limitations.
Although MCD-based methods provide an efficient approximation for BayesNNs,
their predictive uncertainty and calibration ability have been demonstrated to be worse than deep ensembles~\cite{ovadia2019can}.
The introduction of MCD layers after each convolution in vanilla MCD-based BayesNNs can hamper their predictive power, worsening both its accuracy and its uncertainty quantification \cite{bayessegnet}.

In contrast, the multi-exit approach can be interpreted as deep ensembles with a shared backbone network~\cite{ee}, but it lacks flexibility when the calibration requires more predictive outputs than the number of exits.
To alleviate the drawbacks of both of these individual approaches,
we propose multi-exit MCD-based BayesNNs.

\begin{figure}[htb]
\centering
\includegraphics[width=0.49\textwidth]{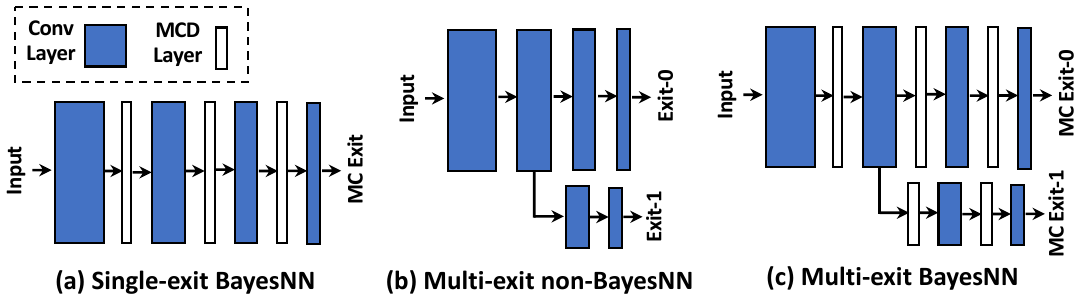}
\vspace{-3.8mm}
\caption{Difference between a single-exit BayesNN, a multi-exit NN and a multi-ext BayesNN.}
\label{fig:multi-exit}
\end{figure}

% Combination
\figref{fig:multi-exit} illustrates the difference between a vanilla MCD-based Bayesian NN, a multi-exit non-BayesNN, and a multi-exit MCD-based BayesNN.
By introducing MCD layers in the multi-exit architecture, the proposed approach can achieve a similar level of calibration ability to deep ensembles.
Moreover, multi-exit MCD-based BayesNNs can perform MC sampling by running the introduced MCD layers multiple times, which enables the ability to generate arbitrary numbers of MC samples.
As discussed above, the usage of MCD layers after every convolution can introduce too much regularization into the network, leading to worse accuracy and worse uncertainty quantification~\cite{qendro}. 
% However, strategic placement of fewer dropout layers can improve performance~\cite{qendro}. 
Also, by placing MCD layers as close to each exit as possible, fewer computations are required since the non-Bayesian results can be cached and reused for different MC samples. 
Therefore, rather than adopting the fully MCD-based approach, 
we insert MCD layers starting from exits towards the input.
The number of MCD layers is a hyper-parameter for optimization.
We refer the layers without MCD applied as the non-Bayesian component.

To demonstrate that the multi-exit MCD-based BayesNN with both Bayesian and non-Bayesian components is still a mathematically valid approximation to BayesNNs,
one can interpret the non-Bayesian component as a feature extractor.
Therefore, given an $M$-exit architecture with inputs $\mathbf{X}$, our approach first maps the data from input space into feature space by using $f_{i}(X)$, where $f_{i}(.)$ denotes the feature extractor of each exit with $1 \leq i \leq M$.
By replacing the $\mathbf{X}$ with $f_{i}(\mathbf{X})$ in the mathematical proof provided by~\cite{gal2016dropout},
our multi-exit MCD-based BayesNN can be interpreted as the ensembles of approximated BayesNNs built upon the feature space.

Another benefit of multi-exit BayesNNs is the lower computational cost of generating MC samples than single-exit BayesNNs.
Given that the floating-point operations (FLOPs) of the main body and all the exits are respectively $FLOP_{main}$ and $FLOP_{exit}$.
As getting one MC sample needs to run the entire network in single-exit BayesNNs, the computational cost of running $N_{sample}$ MC samples can be formulated as:
\begin{equation}\label{eq:cost_single_exit}
    N_{sample} \times (FLOP_{main} + FLOP_{exit}).
\end{equation}
In contrast,
the required FLOPs of an $N_{exit}$ multi-exit BayesNN to get the same number of MC samples is:
\begin{equation}\label{eq:cost_multi_exit}
    FLOP_{main}+ \frac{N_{sample}}{N_{exit}} \times FLOP_{exit}.
\end{equation}
The reduction rate is given by dividing Equation~\ref{eq:cost_single_exit} by Equation~\ref{eq:cost_multi_exit},

\begin{equation}
    \frac{1 + \alpha}{\frac{1}{N_{sample}} +  \frac{\alpha}{N_{exit}}},
\end{equation}
where $\alpha = \frac{FLOP_{exit}}{FLOP_{main}}$.
The reduction rate varies by different multi-exit architectures, depending on $N_{sample}$, $N_{exit}$ and $\alpha$.

Section \ref{sec:multiexit} discusses the wide variety of possible methods in which multi-exit networks can be created and trained. In this work, the exit branches are placed according to the approach used in~\cite{bidistillation}. Semantic groupings are formed for each network, splitting the network architecture into ``blocks" separated by pooling layers. An exit branch is then placed after each of these blocks. In order to allow for more direct validation of the work performed in this paper, the bidirectional distillation training method in~\cite{bidistillation} is used.

\section{Transformation Framework}\label{sec:framework}
\subsection{Framework Overview}

\begin{figure}[htb]
\centering
\includegraphics[width=0.49\textwidth]{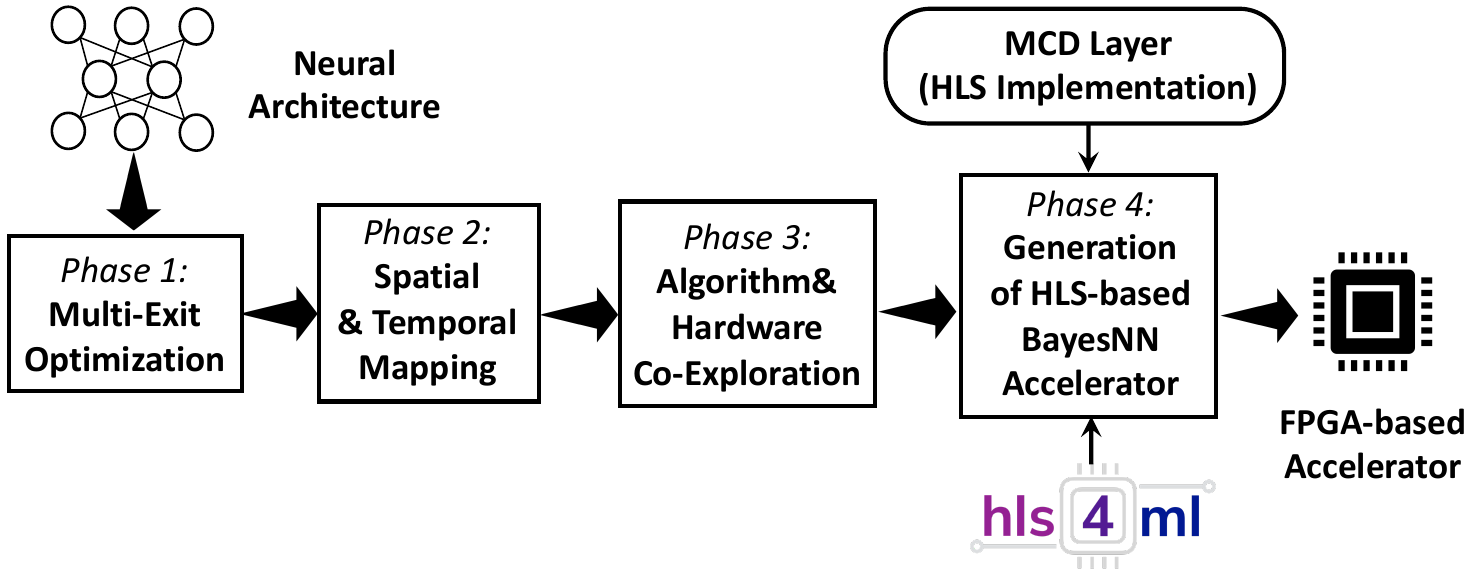}
\vspace{-7mm}
\caption{Framework Overview.}
\label{fig:tool_overview}
\end{figure}

An overview of our transformation framework is presented in~\figref{fig:tool_overview}.
% To generate FPGA-based multi-exit BayesNN accelerator,
There are four phases in our proposed framework: \textit{(1)} construction and optimization of multi-exit MCD-based BayesNNs, \textit{(2)} spatial and temporal mapping optimization, \textit{(3)} design space exploration for both algorithm and hardware design and \textit{(4)} generation
of BayesNN accelerators based on HLS (High-Level Synthesis). 

Given the neural architecture of a non-BayesNN as the input,
the first phase constructs a multi-exit MCD-based BayesNN by optimizing the multi-exit architecture, the number of MCD layers, and dropout rates.
The second phase applies both temporal and spatial mappings to improve the hardware performance.
% An algorithm and hardware  co-exploration is performed in the third stage to optimize several design parameters such as the bitwidth and execution strategy.
The third phase involves algorithm and hardware co-exploration to optimize design parameters such as bitwidth and execution strategy.
The last phase generates the corresponding HLS-based hardware accelerator.
We adopt the design flow and HLS template of non-Bayesian layers from \textit{HLS4ML}.
To support the generation of BayesNN accelerators, we  add the HLS-based implementation of MCD layers and \textit{Keras}-to-HLS conversion into the design flow.

\subsection{Multi-exit Optimization: Phase 1}

\begin{figure}[htb]
\centering
\includegraphics[width=0.4\textwidth]{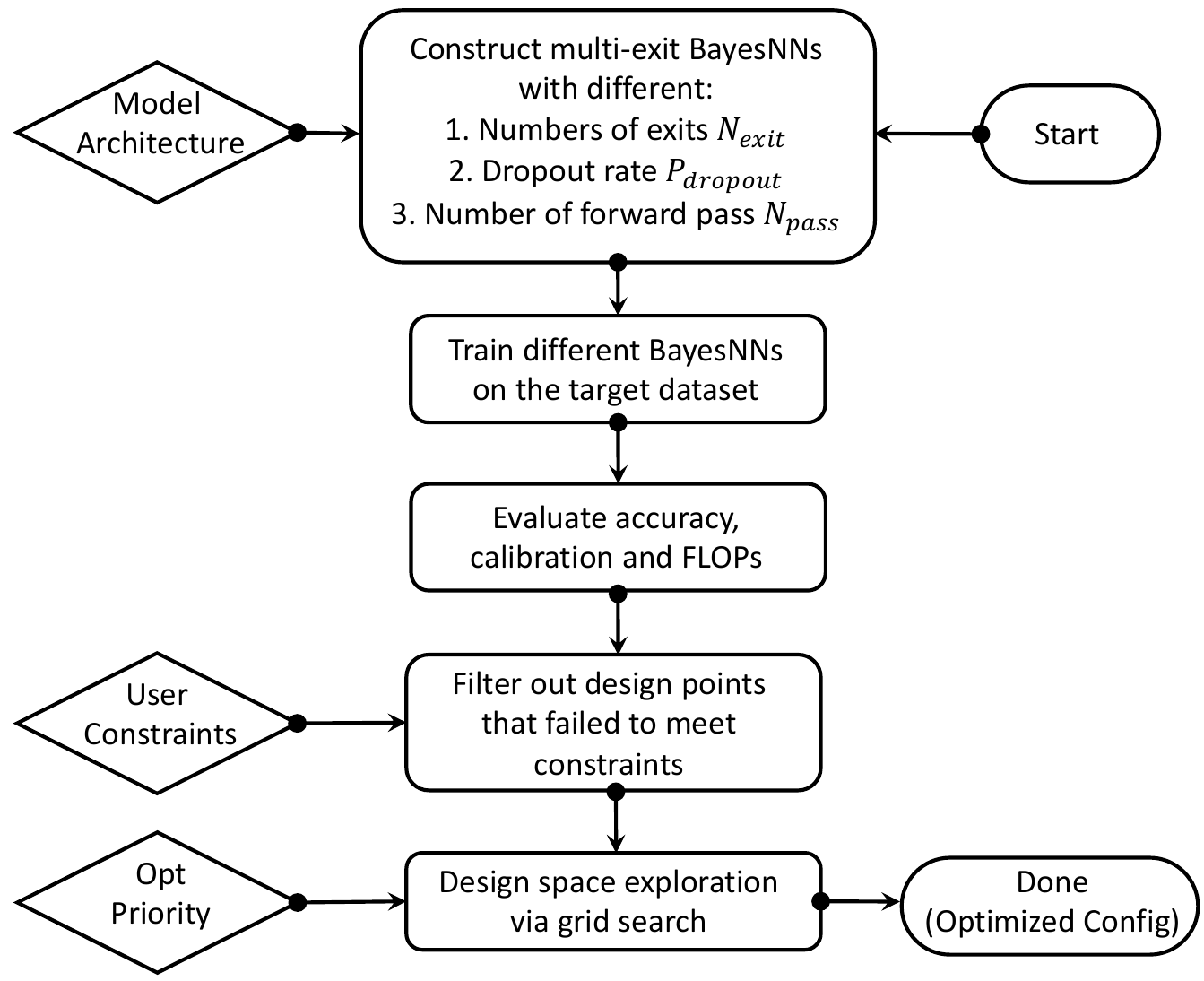}
\caption{Optimization flow.}
\label{fig:flow_chart}
\end{figure}

Since the number of exits is a design parameter in multi-exit BayesNNs, it presents a larger design space than the conventional BayesNNs.
% Assuming there are $N_{exit}$ in a multi-exit BayesNN, 
Given that a multi-exit BayesNN has $N_{exit}$ exits,
the number of forward passes $N_{pass}$ required to produce the total number of MC samples $N_{sample}$ is given by
$N_{pass} = \frac{N_{sample}}{N_{exit}}$.
The higher number of $N_{exit}$ and $N_{pass}$ may improve both the accuracy and calibration. However,
it can degrade hardware performance due to the larger amount of computational and memory demands. 
As different applications and tasks may have different requirements for algorithm and hardware performance,
we propose an optimization exploration flow as shown in~\figref{fig:flow_chart}.
% to explore the design space under different scenarios.

The optimization flow starts from the model construction of multi-exit BayesNNs given the input model architecture.
By inserting $N_{exit}$ exits and an MCD layer with dropout rate $P_{dropout}$, different multi-exit BayesNNs are constructed and trained on the target dataset.
When the training finishes,
we evaluate different metrics for multi-exit BayesNNs, including accuracy, calibration and the amount of floating-point operations (FLOPs).
% -- note that this does not mean floating-point operations per second, and lower amount of FLOPs are better.

Based on the evaluated performance, the design points that do not meet user constraints are filtered out.
Then,
according to the optimization priority,
design space exploration is performed to find the optimal design configuration via grid search.
% The optimization priority can be one of accuracy, uncertainty and the amount of FLOPs.
The optimization priority can be based on accuracy, calibration and the amount of FLOPs.
The final optimized design is fed into the next stage for hardware design generation.

\subsection{Spatial and Temporal Mappings: Phase 2}\label{subsec:spt_temp_map}
\begin{figure}[htb]
\centering
\vspace{-2.0mm}
\includegraphics[width=0.49\textwidth]{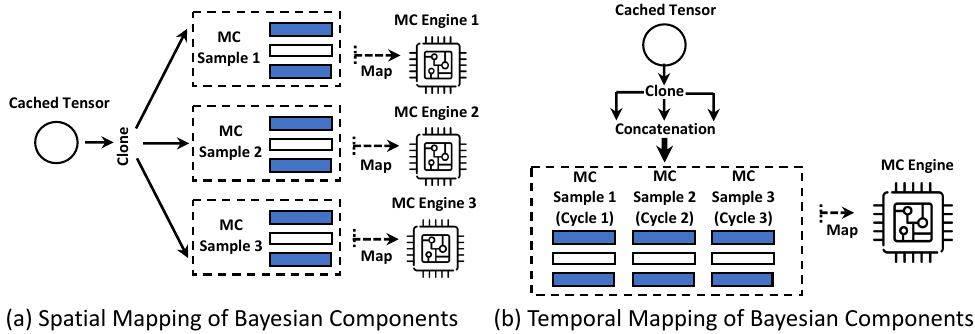}
\vspace{-5.0mm}
\caption{Spatial and temporal mappings for Bayesian components.}
\vspace{-1.0mm}
\label{fig:spt_temp_map}
\end{figure}

The inference of Bayesian components requires multiple forward passes to obtain different MC samples.
This Bayesian-related computation exhibits concurrency along the sampling dimension compared with conventional non-Bayesian NNs, enabling new parallelism strategies in hardware design.
Therefore, we propose two mapping strategies, spatial and temporal mappings, to accelerate the Bayesian component of Multi-exit MCD-based BayesNNs, which are illustrated in~\figref{fig:spt_temp_map}.

In both mapping strategies, the tensor generated from the last non-Bayesian layer is cached and cloned into multiple copies.
As shown in~\figref{fig:spt_temp_map}(a), the spatial mapping deploys separate hardware MC Engines for different MC samples.
Although spatial mapping effectively reduces latency by spatially parallelizing the sampling dimension, it also significantly increases resource use when the number of MC samples becomes large.
To alleviate this issue,
we propose temporal mapping that shares one MC Engine among multiple MC samples.
As shown in~\figref{fig:spt_temp_map}(b),
the cloned copies are concatenated before feeding into the shared engine, which maps different MC samples one by one onto a single MC Engine. 
% In this work, we use a mix of spatial and temporal mappings to satisfy different latency and resource constraints.
Our approach optimizes the mix of spatial and temporal mappings to meet different latency and resource constraints.

\subsection{Algorithm and Hardware Co-Exploration: Phase 3}
Our hardware accelerator contains different design parameters, such as
the implementation strategy used in \textit{HLS4ML}, the reuse factor specified for each layer, and the mapping strategy adopted for the Bayesian component.
On the algorithmic side, given the input model architecture,
there are several hyper-parameters that can be optimized, including the channel number and the bitwidth of activations and weights.
We adopt grid search to optimize both algorithm and hardware design parameters with the requirement of not reducing the algorithmic performance compared to the default configurations.
% To avoid vast design space and expensive search costs,
To reduce search costs, we experiment with heuristics such that
the bitwidth is chosen from $\{4, 6, 8, 16\}$,
and the channel number is selected from $\{C, \frac{C}{2}, \frac{C}{4}, \frac{C}{8}\}$ with $C$ being the original number of channels. Users can also define other search space.

\subsection{Generation of FPGA-based Accelerator: Phase 4}

The generation of hardware accelerators is based on \textit{HLS4ML} and our customized MCD design template. 
The generated HLS-based BayesNN accelerators can then be fed into Vivado-HLS for synthesis and implementation to get the final bitstream for onboard testing.
The pseudocode of HLS-based implementation of MCD is presented in Algorithm~\ref{algrm:hls_mcd}.
The HLS directive \textit{HLS PIPELINE} is used to improve the overall performance.
We cache the temporary result in the variable \textit{temp}, before generating the final outputs.
% The corresponding hardware design is presented in~\figref{fig:dropout}.
The hardware design receives the stream input data from the preceding layer, and produces stream outputs to the following layer.
The dropout rate $P_{dropout}$ is a design parameter specified by the user at the beginning of running each model.
A multiplexer is used to select either $0$ or the result of the multiplication between inputs and dropout rate $P_{dropout}$.
The control signal of the multiplexer is generated by comparing $P_{dropout}$ with \textit{uniform\_random}.
To support the MCD layer with arbitrary $P_{dropout}$, a random number generator is used in our design to generate \textit{uniform\_random}.

\begin{algorithm}
\caption{Pseudocode of MCD layer}\label{algrm:hls_mcd}
  \begin{algorithmic}[1]
   \State \textbf{Input}: input[$dropout\_size$], keep\_rate
   \State \textbf{Output}: output[$dropout\_size$]
    \For{($i$ from $0$ to $dropout\_size$)} \Comment{\#pragma PIPELINE}
        \State temp = input[i]
        \State uniform\_random = random\_number\_generator()
        \If{(uniform\_random $>$ keep\_rate)} temp = 0
       \EndIf
       \State output[i] = temp * keep\_rate
    \EndFor
  \end{algorithmic}
\end{algorithm}

% \begin{figure}[htb]
% \centering
% \includegraphics[width=0.39\textwidth]{figures/hw_dropout.pdf}
% \caption{Hardware design of MCD.}
% \label{fig:dropout}
% \end{figure}

\section{Experiments and Evaluation}
Our optimization framework is implemented in Python $3.8.12$, PyTorch $1.11.0$, and Keras $2.9.0$.
We use Vivado-HLS $2020.1$ for hardware implementation.
% design, and  for synthesis and place\&route.
QKeras is used for quantization.
% \textcolor{red}{We use QKeras is used to apply the quantization in the design}
% A server with $6$ Nivida GPUs is used for training.
The latency and resource consumption are obtained from C-synthesis reports provided by Vivado-HLS.
Vivado $2020.1$ is used to run place and route for the final designs.
We set Xilinx Kintex XCKU115 as our target FPGA board.
All the designs are optimized by our spatial-temporal mapping and algorithm-hardware co-exploration to ensure they can be fitted into the target platform.

\subsection{Cost of Being Bayesian}

\begin{figure}[htb]
\centering
\includegraphics[width=0.49\textwidth]{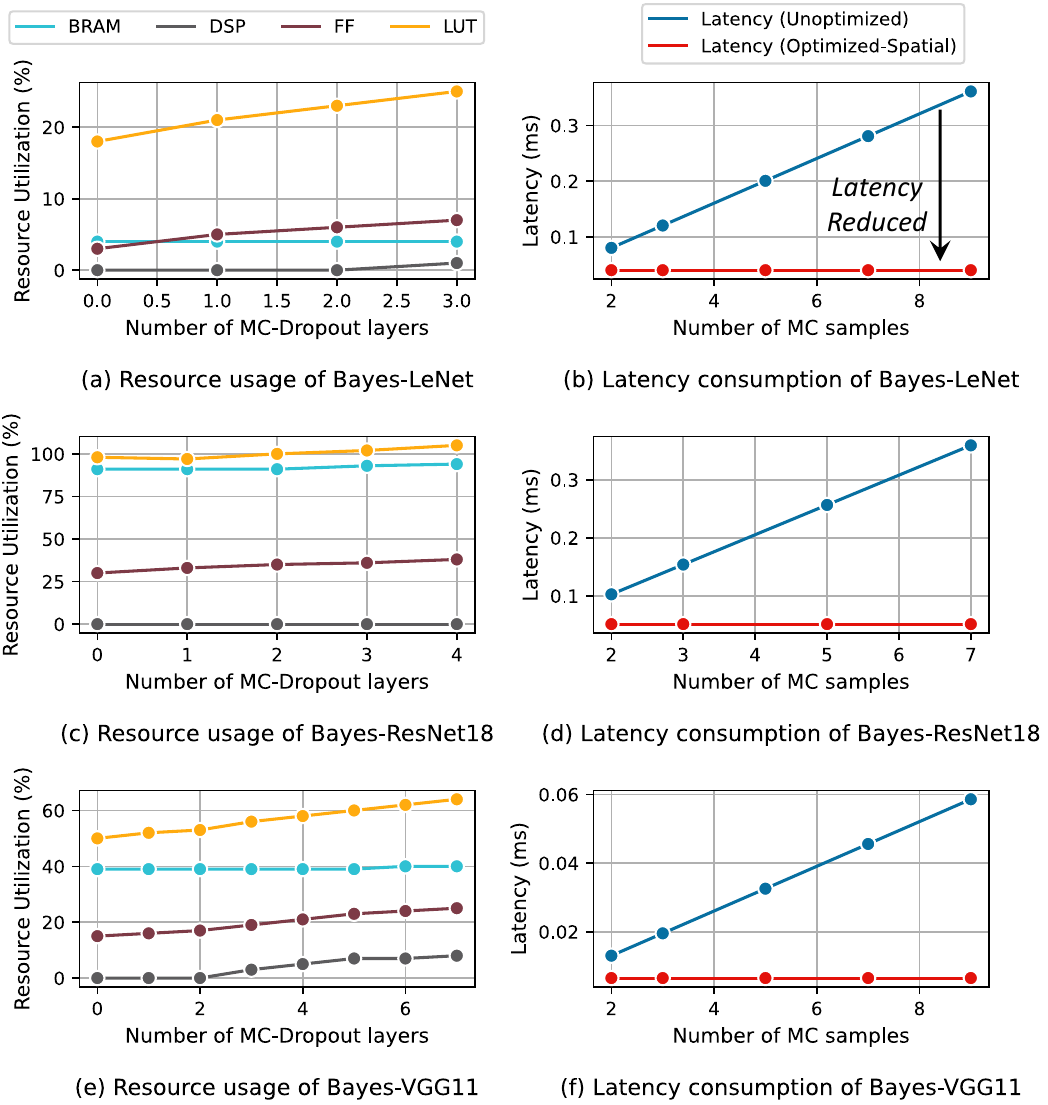}
\vspace{-3mm}
\caption{{Resource consumption and latency of Bayes-LeNet, Bayes-ResNet18 and Bayes-VGG11 with quantization and custom number of channels.}}
\vspace{-5mm}
\label{fig:lat_cost}
\end{figure}

% Previous experiments have demonstrated the benefits of multi-exit MCD-based BayesNNs in terms of accuracy and uncertainty prediction.
% In this experiment, we evaluate the hardware cost of supporting multi-exit MCD-based BayesNNs on FPGA.
The first experiment evaluates the hardware cost of supporting BayesNNs on FPGA.
Compared with non-BayesNN designs, 
the hardware overhead of BayesNN accelerators comes from the use of MCD layers and the need to run multiple MC samples.
To quantitatively investigate the cost,
we evaluate the three BayesNNs on different datasets, i.e., \textit{LetNet5} on MNIST, \textit{ResNet-18} on CIFAR-10, and \textit{VGG-11} on SVHN.
As this experiment aims to evaluate the cost of being Bayesian,
we use one exit on each model to eliminate the hardware overhead introduced by the multi-exit optimization.
The custom configurations of these models, including quantization channel settings, are available in our open-sourced code.
% To fit models on our target FPGA board,
% we apply QKeras quantization and adopt smaller channel numbers.
The results are presented in~\figref{fig:lat_cost}.
% presents the results that are obtained from C-synthesis.

% While evaluating the resource overhead,
% we generate and synthesis the HLS designs 
To evaluate the resource overhead, we generate and synthesize the designs using temporal mapping~(\secref{subsec:spt_temp_map}) with different numbers of MCD layers.
The resource consumption of Block RAM (BRAM), DSP, Flip-Flop (FF) and LUT is shown on the left of~\figref{fig:lat_cost}.
The utilization of logic resources, including FF and LUT, shows an increasing trend when the number of MCD layers becomes larger.
The increase of DSP is not significant, except for \textit{Bayes-VGG11} which has an $8\%$ increase with $7$ MCD layers.
% However, the overhead of both FF and LUT consumption is still within $8$\% across three different models while increasing the number of MCD layers.
As the design of the MCD layer does not require BRAM in the design,
the BRAM consumption remains the same across different numbers of MCD layers on all three models.
% Given that three or four layers of MCD can achieve the best performance trade-off between accuracy, ECE and FLOPs as indicated in~\secref{subsec:eff_memc},
% the cost of enabling uncertainty prediction on DNNs accelerators is within $5$\%.
To measure the latency cost of MC sampling,
we evaluate the designs with one MCD layer using different numbers of MC samples.
To demonstrate the effect of spatial mapping optimization,
we compare the latency of two implementations with and without spatial mapping.
For the unoptimized version,
we assume a single engine is used for multiple MC samples.
% For the design with temporal mapping,
% an increase can be observed in its latency consumption when the number of MC samples becomes large. 
% As we can see from the right of~\figref{fig:lat_cost},
% an increase in latency consumption can be observed when the number of MC samples becomes large 
The right of~\figref{fig:lat_cost} shows an increase in latency when the number of MC samples becomes large in the unoptimized implementations.
In contrast, due to the parallelization of different MC samples,
the latency cost of spatial mapping almost stays the same with the increasing number of MC samples, demonstrating the effectiveness of spatial mapping.

% When more MC-Dropout layers are added into the design,
% there is an increasing trend in all three models.
% Especially in the BayesNNs with small numbers of channels, such as \textit{Bayes-LeNet5} and \textit{Bayes-ResNet18},
% the latency consumption of MC-dropout layers is lower than $0.5$ms.

\begin{table*}[tb]
\centering
\caption{Performance comparison among SE CNNs, MCD BayesNNs, ME and MCD-ME BayesNNs with 32-bit floating point (FP32).}
\vspace{-2.5mm}
\vspace{0.3cm}
\label{tb:multi-exit}
\setlength\tabcolsep{1pt}
\scalebox{0.9}{
\begin{tabular}{C{2.2cm}C{2.0cm}C{1.8cm}C{2.0cm}C{2.0cm}C{2.0cm}C{1.8cm}C{2.0cm}C{1.8cm}}
\toprule
& \multicolumn{4}{c}{ResNet18 (FP32)} & \multicolumn{4}{c}{VGG19 (FP32)}\\
\cmidrule{ 2 - 9 }
& \multicolumn{2}{c|}{Acc-Opt} & \multicolumn{2}{c|}{ECE-Opt} & \multicolumn{2}{c|}{Acc-Opt} & \multicolumn{2}{c}{ECE-Opt} \\
\cmidrule{ 2 - 9 }
 & Accuracy & FLOPs & ECE & FLOPs & Accuracy & FLOPs & ECE & FLOPs \\
\midrule
SE & $0.752 \pm 0.002$ & $\boldsymbol{1.00}$ & $0.0840 \pm 0.0008$ & $1.00$ & $0.693 \pm 0.002$ & $1.00$ & $0.165 \pm 0.006$ & $1.00$\\
\midrule
MCD & $0.758 \pm 0.002$ & $1.00$ & $0.069 \pm 0.001$ & $1.00$ & $0.696 \pm 0.004$ & $1.00$ & $0.131 \pm 0.006$ & $1.00$ \\
\midrule
ME & $0.7719 \pm 0.0006$ & $1.026 \pm 0.003$ & $0.017 \pm 0.002$ & $1.026 \pm 0.003$ & $0.747 \pm 0.002$ & $\boldsymbol{0.977}$ & $0.025 \pm 0.001$ & $0.46 \pm 0.05$\\
\midrule
MCD+ME (Ours) & {$\boldsymbol{0.776 \pm 0.001}$} & $1.019 \pm 0.004$ & $\boldsymbol{0.014 \pm 0.001}$ & $\boldsymbol{0.672 \pm 0.003}$ & $\boldsymbol{0.747 \pm 0.001}$ & $0.982$ & $\boldsymbol{0.017 \pm 0.001}$ & $\boldsymbol{0.45 \pm 0.02}$\\
\bottomrule
\end{tabular}
}
\end{table*}

\begin{table*}[t]
\centering
\caption{Performance comparison of our final FPGA designs with CPU, GPU, and other FPGA-based implementations. \vspace{0.1cm}}
\label{tb:compare_sota}
\setlength\tabcolsep{1pt}
\scalebox{0.88}{
\begin{tabular}{C{3.99cm}C{2.5cm}C{2.5cm}C{2.5cm}C{2.0cm}C{2.0cm}C{2.0cm}C{2.0cm}}
\toprule
{}&{\bf CPU}&{\bf GPU}&{\bf ASPLOS'18~\cite{cai2018vibnn}} &{\bf \bf DATE'20~\cite{awano2020bynqnet}}& {\bf DAC'21~\cite{fan2021high}}&{\bf TPDS'22~\cite{fan2022accelerating}}& {\bf Our Work}\\
\midrule 
{\bf Platform}& {Intel Core i9-9900K} &{NVIDIA RTX 2080 } &{Altera Cyclone V} & {Zynq XC7Z020}&{Arria 10 GX1150}&{Arria 10 GX1150}  &{XCKU 115} \\
\midrule
{{\bf Frequency} (MHz)} & 3600& 1545& 213 & 200 &{225}& {220}&{\SI{181}{}}\\\midrule
{{\bf Technology}} & {14 nm} & {12 nm} & {28 nm}  & {28 nm} &{20 nm}& {20 nm}&{20 nm}\\\midrule
{{\bf Power} (W)} & {205}& {236}& {6.11} & {2.76} &{45.00}& {43.6}&{4.6}\\\midrule
% {\bf Model} & {\textit{Bayes-LeNet5}}& {{\textit{Bayes-LeNet5}}}&  {\textit{Bayes-FC}}& {\textit{Bayes-FC}}&{\textit{Bayes-LeNet5}}&
% {\textit{Bayes-LeNet5}}& {{\textit{Bayes-LeNet5}}}\\\midrule
{\bf Latency} (ms)& {1.26}& {0.57}& {5.5}& {4.5}& {0.42}& {0.32}&{0.89}\\\midrule
{\bf Energy Efficiency} (J/Image) & {0.258}& {0.134}& {0.033}& {0.012}& {0.019}& {0.014}&{0.004}\\\bottomrule
\end{tabular}}
\vspace{0.0mm}
\end{table*}

\subsection{Effect of Multi-Exit BayesNNs}\label{subsec:eff_memc}
To demonstrate the advantage of multi-exit BayesNNs over the baseline approaches,
we evaluate two commonly-used multi-exit models, \textit{VGG19} and \textit{ResNet18} for image classification.
Cifar100 dataset, a curated subset of a larger dataset scraped from the web containing photo-realistic tiny $32 \times 32$ images with a single main object, is used in this experiment.
% We use Expected Calibration Error (ECE)~\cite{ovadia2019can} as a metric evaluate predictive uncertatinty.

We compare four different implementations: \textit{i)} Single-exit model with only one exit at the end of the network (SE). There is no MCD or Multi-Exit applied, which is the original implementation of both the  \textit{ResNet-18} and \textit{VGG-19}. \textit{ii)} MCD-based BayesNN without multi-exit (MCD). The MCD is only applied to the single exit of the network.
\textit{iii)} Multi-exit model without MCD (ME). We add one exit after each ResNet and VGG block to make multiple exits.
\textit{iv)} MCD-based BayesNN with multi-exit (MCD + ME). The MCD is applied to every exit of the network.
Stochastic gradient descent (SGD) is used with a weight decay of $5 \times 10^{-4}$, an initial learning rate of $0.1$ and a momentum of $0.9$, along with a batch size of $64$. 

As discussed previously, the usage of too many dropout layers in a BayesNN can overregularize the network and adversely affect performance. 
% However, there is no rule of thumb to find the best balance between the level of dropout and calibration. 
However, there is no standard method to find the best balance
between the level of dropout and calibration.
Therefore, a small grid search is performed over the following dropout rates: 0.125, 0.25, 0.375, 0.5. Similarly, the confidence threshold which optimally balances the computational cost and the network performance is found through testing the same thresholds as in \cite{sdn}: 0.1, 0.15, 0.25, 0.5, 0.6, 0.7, 0.8, 0.9, 0.95, 0.99, 0.999. It is noted that two sets of results from performing confidence-based exiting are calculated, using the predictions at each exit or the largest possible ensemble at each exit respectively. Each ensemble is an equally weighted average of the predictions from each exit, as in \cite{ee}.

The grid search covers all combinations of the above two parameters, which is applied to the applicable networks. The predictions from each of the exits and the ensembles formed by averaging the results from each exit are calculated, alongside the predictions from confidence exiting. The best results are presented in~\tabref{tb:multi-exit} with the calibration captured by expected calibration error (ECE)~\cite{fan2022accelerating}: a low value of ECE denotes a higher quality.
As the dropout rate of MCD and the confidence threshold of multi-exit may affect both accuracy and calibration,
two configurations for each implementation and model are reported: those that achieve the highest accuracy (\textit{Acc-Opt}) and those with the lowest ECE (\textit{Acc-ECE}).
For each configuration, we also calculate the FLOPs as a fraction of the SE implementation.

On \textit{ResNet18},
our approach, MCD + ME, improves the accuracy by $2.4$\% $\pm 0.002$\% with only $0.019$ times more FLOPs compared with the SE implementation.
Our method also shows higher accuracy than both MCD and ME implementations.
In \textit{Acc-ECE},
we achieve the lowest ECE and FLOPs among four implementations.
A similar trend can also be observed in \textit{VGG-19}. Moreover, our approach can match or outperform both of the methods individually, while costing a similar amount of FLOPs. The best model is able to massively reduce the ECE of the SE implementation by $0.148 \pm 0.006$, an improvement of almost 90\%, while costing less than half the amount of FLOPs. 
These results show that multi-exit BayesNNs can lead to better calibrated and more powerful networks, while costing similar or fewer FLOPs.

\subsection{Comparison with CPU, GPU, and FPGA implementations}
To demonstrate the energy efficiency of our approach,
we also compare it against CPU, GPU, and other FPGA-based implementations.
The comparison uses MNIST dataset since it is the most common dataset across different work~\cite{cai2018vibnn, awano2020bynqnet, fan2021high, fan2022accelerating}.
As both~\cite{cai2018vibnn} and~\cite{awano2020bynqnet} do not support \textit{Bayes-LeNet5},
we use their reported throughput (GOP/s) to estimate their performance on \textit{Bayes-LeNet5}.
The performance is obtained by using three MC samples.
Both CPU and GPU performance are quoted from the vanilla implementations of MCD-based BayesNNs in~\cite{fan2022accelerating}.
Although there are some other BayesNN accelerators~\cite{wan2020fast,wan2021shift},
they do not report any end-to-end latency and energy consumption.

As shown in~\tabref{tb:compare_sota},
our design achieves $65$ and  $33$ times higher energy efficiency than CPU and GPU implementations, despite the FPGA adopting 20nm technology while the CPU adopting 14nm technology and the GPU adopting 12nm technology.
Our accelerator also shows lower latency and better energy efficiency than both~\cite{cai2018vibnn} and~\cite{awano2020bynqnet}.
Although both~\cite{fan2021high} and~\cite{fan2022accelerating} are faster than our design,
they consume much higher energy due to the high resource utilization and frequent data transfer between on-chip and off-chip memory, leading to nearly $5$ and $4$ times lower energy efficiency than our design.
Also, compared with their Verilog-based implementations, our HLS-based accelerator has advantages in development time~\cite{pelcat2016design}, which can improve designer productivity and can facilitate extending our approach to cover other NNs such as LSTM~\cite{hochreiter1997long}.
\tabref{tb:power_breakdown} provides the power consumption breakdown obtained from the Xilinx Power Estimator (XPE) tool after place and route.
The dynamic power occupies $72\%$ of the total power.
The logic\&signal and IO consume most of the dynamic power,
accounting for $30\%$ and  $21\%$, respectively.
The high IO power consumption results from our spatial mapping strategy with multiple MC engines running in parallel.

\begin{table}[t]
\centering
\vspace{-1.0mm}
\caption{Power breakdown of our FPGA-based accelerator.}
\label{tb:power_breakdown}
\setlength\tabcolsep{1pt} 
\scalebox{0.9}{
\begin{tabular}{C{1.7cm}|C{1.1cm}|C{1.1cm}|C{1.0cm}|C{0.8cm}|C{0.8cm}|C{1.2cm}|C{1.2cm}}
\toprule
&\multicolumn{5}{c|}{\textbf{Dynamic} (W)} & \multirow{3}{*}{\textbf{Static}} & \multirow{3}{*}{\textbf{Total}} \\ \cmidrule{2-6}
& \multirow{2}{*}{Clocking} &  {Logic\&} & \multirow{2}{*}{BRAM} &  \multirow{2}{*}{IO} &\multirow{2}{*}{DSP} & \\ 
& & {Signal} & & &  \\ \midrule
\textbf{Used} & 0.374  & 1.359  & 0.422 & 0.998 & 0.191 & 1.299 & 4.6 \\ \cmidrule{1-8}
\textbf{Percentage} & 8\% & 30\% & 9\%  & 21\% & 4\%& 28\%& 100\%\\
\bottomrule
\end{tabular}}
\vspace{-1mm}
\end{table} 

% \subsection{Discussion of Results}
% This paper aims to provide a fast prototype tool for BayesNN accelerators.
% As demonstrated in our experiments,
% the \textit{AutoBayes} transformation process only takes a few minutes to generate a HLS-based accelerator for BayesNNs.
% Although there are several other advanced optimizations~\cite{awano2020bynqnet,cai2018vibnn} that can be used to further improve the hardware performance,
% these techniques are orthogonal to our approaches. 
% All our codes and design will be open-source, and we hope this paper provides an easy-to-reproduce baseline design for the research of BayesNN acceleration.

\section{Conclusion}

This paper proposes a novel multi-exit Monte-Carlo Dropout (MCD)-based Bayesian Neural Networks (BayesNNs).
To facilitate its deployment in real-life applications,
a transformation framework is developed to produce FPGA-based accelerators for multi-exit MCD-based BayesNNs.
Several novel hardware optimizations are introduced for performance improvement.
Comprehensive experiments demonstrate that our approach achieves higher algorithmic and energy efficiency than state-of-the-art designs.
In the future, we aim to optimize the design with zero skipping, support attention-based BayesNNs, and include capabilities such as run-time reconfiguration.

\section*{Acknowledgement}{
The support of UK EPSRC grants (UK EPSRC grants EP/L016796/1, EP/N031768/1, EP/P010040/1, EP/V028251/1 and EP/S030069/1) is gratefully acknowledged.
}

\bibliographystyle{IEEEtran}
\bibliography{refs}

\end{document}